\documentclass[conference]{IEEEtran}
\IEEEoverridecommandlockouts
\usepackage{cite}
\usepackage{amsmath,amssymb,amsfonts}
\usepackage{algorithmic}
\usepackage{graphicx}
\usepackage{textcomp}
\usepackage{xcolor}
\usepackage{tabularx}
\usepackage{url}
\usepackage{caption} 
\def\BibTeX{{\rm B\kern-.05em{\sc i\kern-.025em b}\kern-.08em
    T\kern-.1667em\lower.7ex\hbox{E}\kern-.125emX}}
    
\begin{document}

\title{Railway Anomaly detection model using synthetic defect images generated by CycleGAN}

\author{

\IEEEauthorblockN{Takuro Hoshi}
\IEEEauthorblockA{\textit{JR East Information Systems}}
\and
\IEEEauthorblockN{Yohei Baba}
\IEEEauthorblockA{\textit{JR East Information Systems}}
\and
\IEEEauthorblockN{Gaurang Gavai}
\IEEEauthorblockA{\textit{PARC, A Xerox Company}}
}

\maketitle

\begin{abstract}
Although training data is essential for machine learning, railway companies are facing difficulties in gathering adequate images of defective equipment due to their proactive replacement of would be defective equipment.
Nevertheless, proactive replacement is indispensable for safe and undisturbed operation of public transport.
In this research, we have developed a model using CycleGAN to generate artificial images of defective equipment instead of real images.
By adopting these generated images as training data, we verified that these images are indistinguishable from real images and they play a vital role in enhancing the accuracy of the defect detection models.

\end{abstract}


\section{Introduction}

Railway companies have been working on defect detection of railway equipment. However the equipment is replaced early to ensure the smooth operation of the railway with safety as a priority, leading to railway companies facing a common issue of having a lack of images of defective equipment.\\
We developed a defect detection model that is able to identify railway defects by utilizing the images captured by the railway monitoring system\cite{jreMonitoring} installed on the commercial trains of East Japan Railway Company. However, we faced a problem in which the detection accuracy could not be improved due to the lack of training data for a certain type of defects.

\section{Defect detection pipeline for multiple types of track defects using ResNet\cite{he2015deep}}

\begin{figure}[ht]
  \includegraphics[width=\columnwidth]{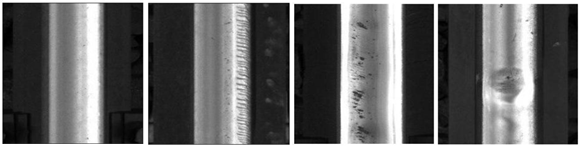}
  \caption{Railway scratches (L to R: normal, scratch, inside-scratch and shelling defects)}
    \label{fig:scratches}
\end{figure}

The sample images of scratch, inside-scratch and shelling defect are shown in Fig.\ref{fig:scratches}. The number of images used for training the defect detection model  are  shown  in  Table\ref{tab:defects}.  It  is  evident  that  the  data  of shelling  defect  is  considerably  less  compared  to  the  scratch and inner side scratch defect.

\begin{table}[ht]
\begin{tabularx}{\columnwidth}{|X|X|X|X|X|}
      \hline
      \textbf{} & \textbf{Normal} & \textbf{Scratch Defect} &  \textbf{Inside-scratch Defect} & \textbf{Shelling Defect}\\
      \hline Training data & 4500 & 992 & 1160 & 82 \\
      \hline Test data & 100 & 50 & 50 & 10 \\
      \hline 
\end{tabularx}
\caption{Number of images used for defect detection model}
\label{tab:defects}
\end{table}

\begin{figure}[ht]
  \centering
  \includegraphics[width=\columnwidth]{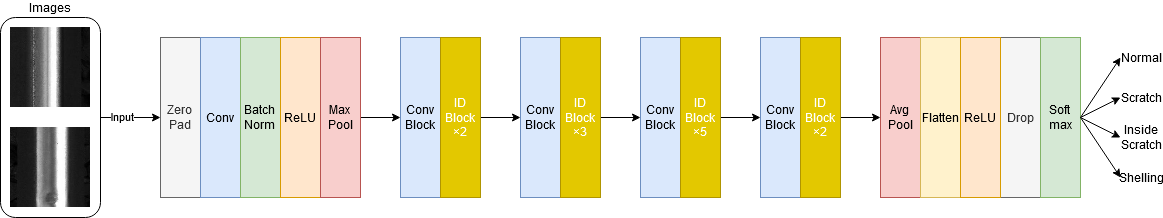}
  \caption{The structure of the anomaly detection model}
    \label{fig:resModel}
\end{figure}

Fine-tuning is a method of relearning the weights of a neural network by replacing the final output layer with an existing model trained in a different domain. The structure of the anomaly detection model is shown in Fig.\ref{fig:resModel}.

\begin{figure}[ht]
  \centering
  \includegraphics[width=\columnwidth]{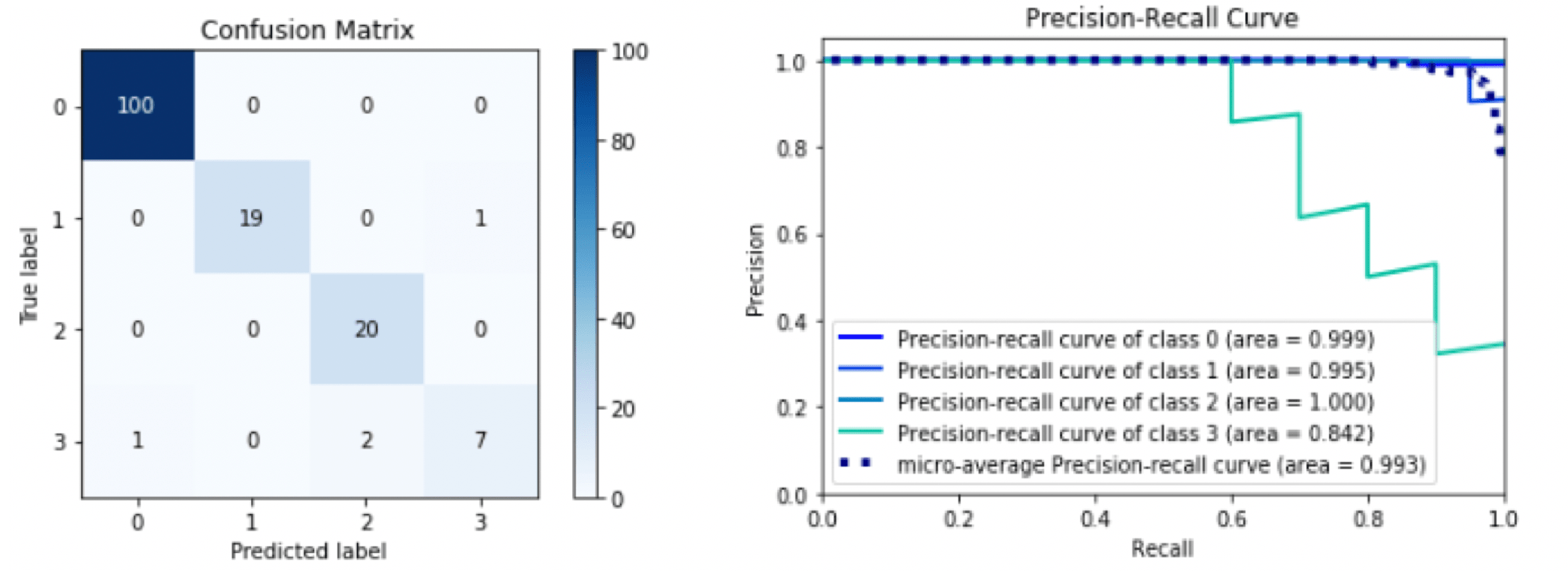}
  \caption{Accuracy of defect detection model\\(0: normal, 1:scratch, 2: inside-scratch, 3: shelling defects)}
    \label{fig:beforeAcc}
\end{figure}

\begin{figure}[ht]
  \centering
  \includegraphics[width=0.25\columnwidth]{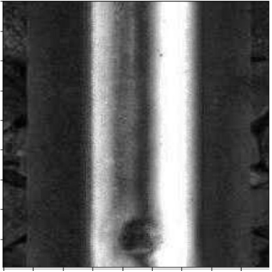}
  \caption{Image of shelling that was not detected}
    \label{fig:shellingImage}
\end{figure}

The defect detection model was unable to identify 1 incident of a superficial shelling defect. The accuracy of the defect detection model is shown in Fig.\ref{fig:beforeAcc} while the image of the shelling that was not detected is shown in Fig.\ref{fig:shellingImage}.

\section{Railway defect generation model using CycleGAN}

\begin{figure}[ht]
  \centering
  \includegraphics[width=\columnwidth]{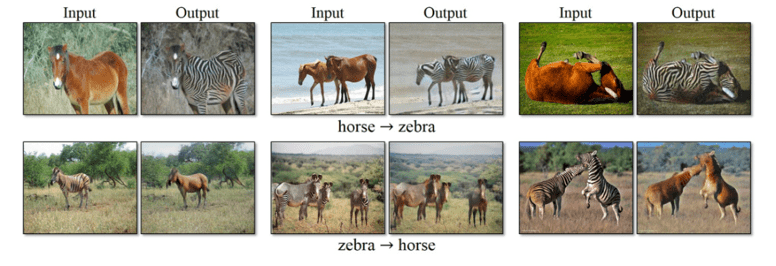}
  \caption{Examples of images generated by CycleGAN\cite{zhu2020unpaired}}
    \label{fig:cycle_examples}
\end{figure}

\begin{figure}[ht]
  \centering
  \includegraphics[width=\columnwidth]{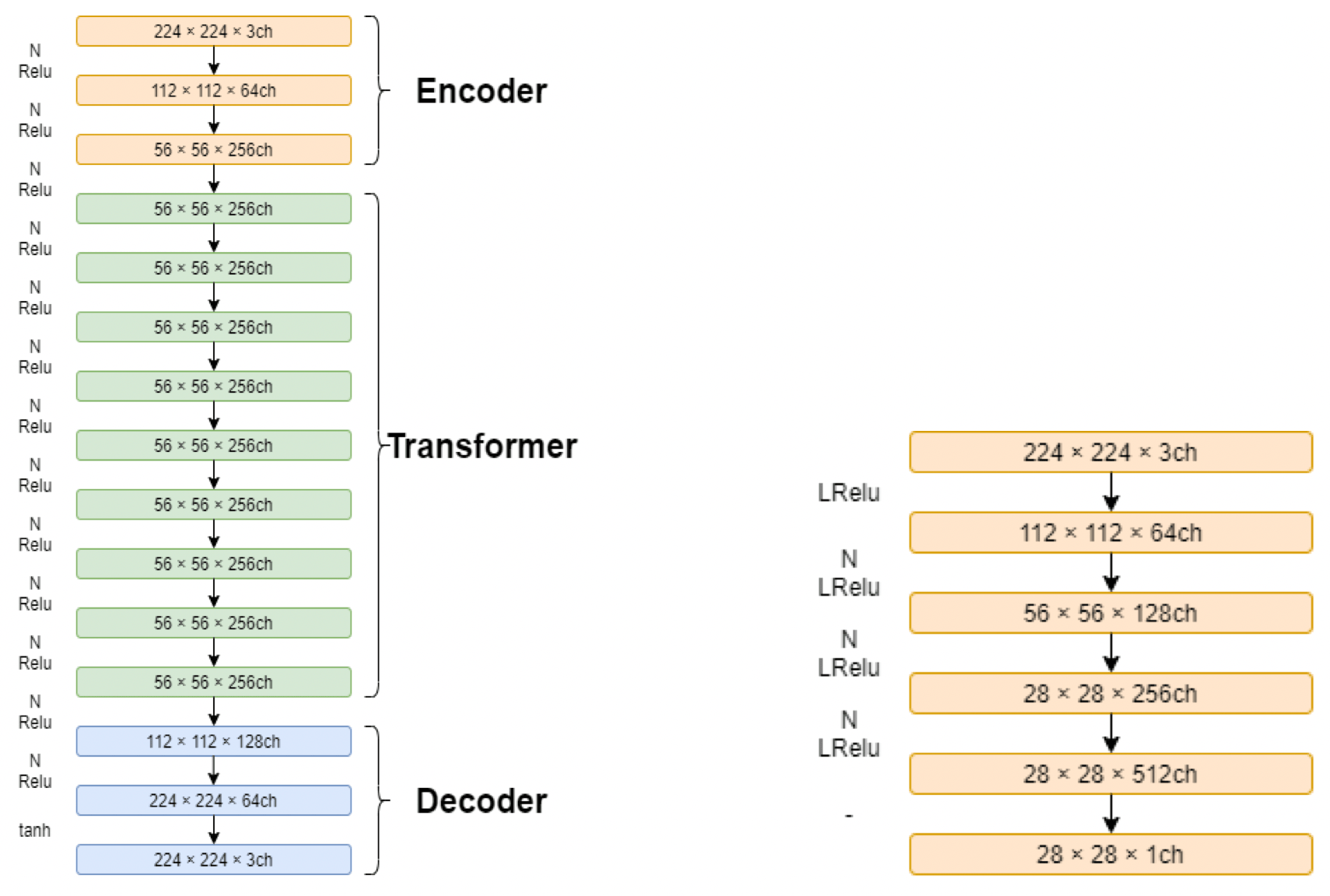}
  \caption{Configuration of Generator and Discriminator}
    \label{fig:cycle_model}
\end{figure}

Due to the lack of adequate training data in order to generate images of railway defects, we have chosen an algorithm known as CycleGAN. CycleGAN is a type of style transfer method where an image is generated based on original input with its characteristics altered. The examples of images generated by CycleGAN and the algorithm used are shown in Fig.\ref{fig:cycle_examples} and the configuration of the Generator and Discriminator used in this research is shown in Fig.\ref{fig:cycle_model}.

\begin{figure}[ht]
  \includegraphics[width=\columnwidth]{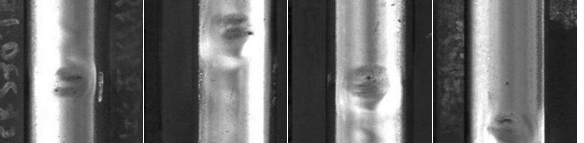}
  \caption{Image conversion using CycleGAN}
    \label{fig:cycle_results}
\end{figure}

\begin{figure}[ht]
  \includegraphics[width=\columnwidth]{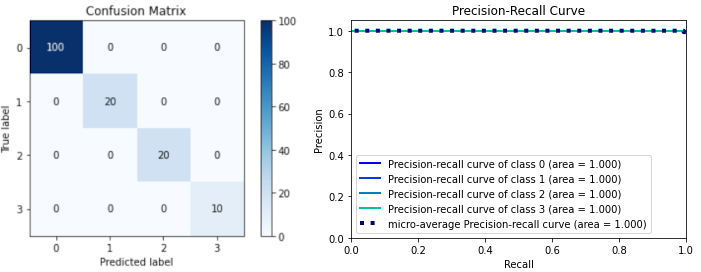}
  \caption{Accuracy of defect detection model after addition of CycleGAN generated images.}
    \label{fig:after_acc}
\end{figure}

The image conversion using CycleGAN model is shown in Fig.\ref{fig:cycle_results}. The images converted by CycleGAN were difficult for human eyes to distinguish from actual images on rails.
From the 1000 images generated by CycleGAN model, 70 of them which were similar to the undetected shelling defect, were chosen and added to the defect detection model as training data. The accuracy of detecting shelling defects was successfully improved and the test data showed that the average area under the curve was 1, higher compared with 0.993 of Fig.\ref{fig:beforeAcc}. The test results are shown in Fig.\ref{fig:after_acc}.

\section{Visualization class activation mapping using CAM\cite{Selvaraju_2019} and t-SNE\cite{vanDerMaaten2008}}

\begin{figure}[ht]
  \centering
  \includegraphics[width=\columnwidth]{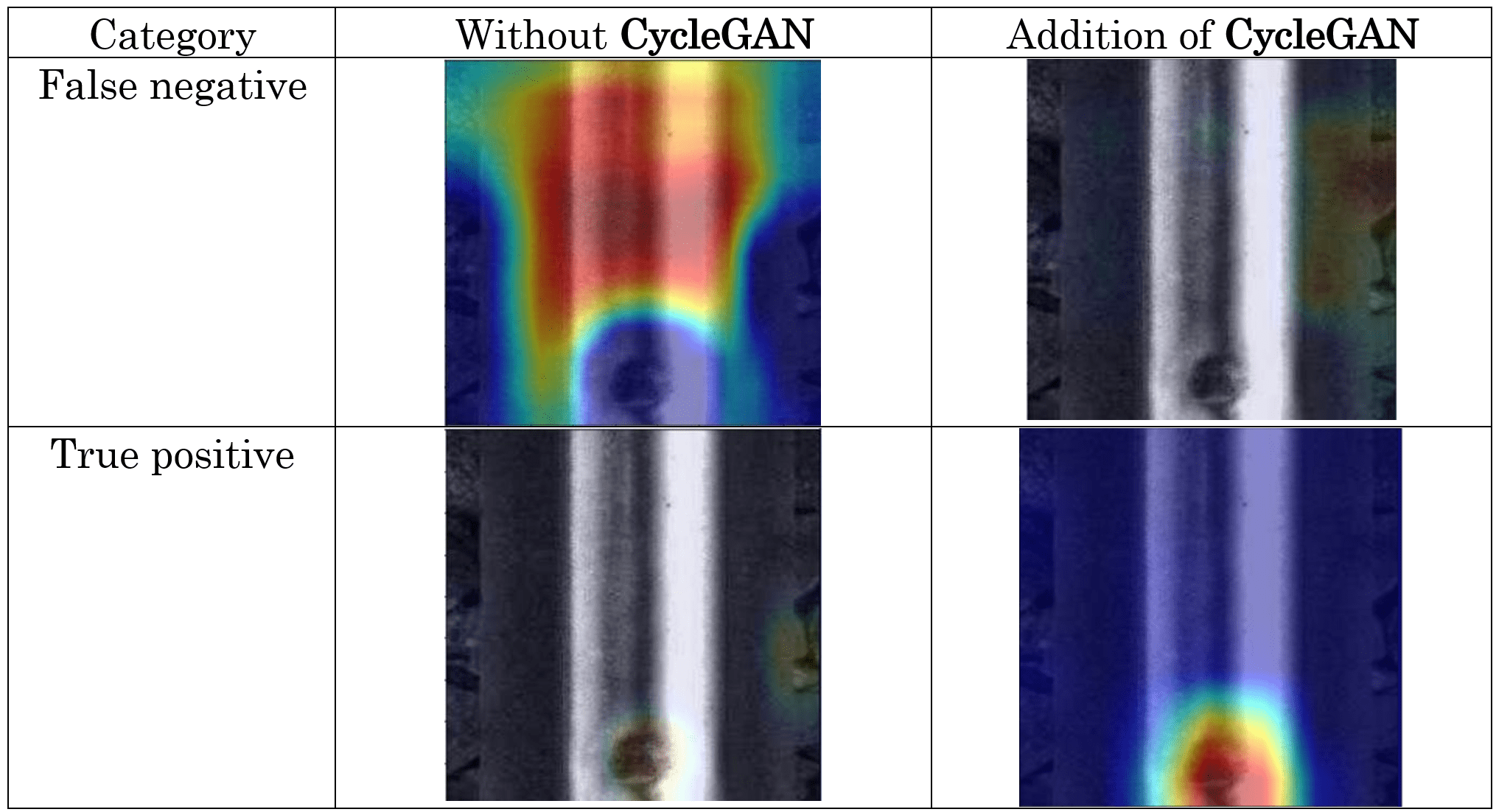}
  \caption{Heat map changes before and after addition of images generated by CycleGAN}
    \label{fig:heatmap}
\end{figure}

In order to visually confirm the changes that occurred following the addition of training data to the defect detection model, we have utilized a method known as Class Activation Model (CAM)  to create a heat map. This is a visual representation of which part of the image the defect detection model focused on when classifying the images. The parts that were used to classify the images are shown in red while the parts that were not are shown in blue. The results are shown in Fig.\ref{fig:heatmap}.

\begin{figure}[ht]
  \centering
  \includegraphics[width=\columnwidth]{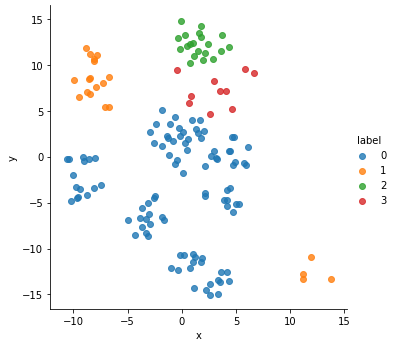}
  \caption{t-SNE graph input of images generated by CycleGAN\\(0: normal, 1:scratch, 2: inside-scratch, 3: shelling defects)}
    \label{fig:afterTsne}
\end{figure}

As shown in Fig.\ref{fig:heatmap}, the defect detection model before the input of images generated by CycleGAN reacted strongly towards the linear parts other than the defects on the rails. It can be thought that small shelling defects were not detected as such due to the lack of data on images of defective equipment. It is evident that adding the generated images of shelling defects as training data enables the neural network to focus on the shelling defects and identify them as such.
t-SNE was used to perform classification analysis (clustering) on the data of 2048-dimensions immediately before the generation of final results by the defect detection model. The results are shown in Fig.\ref{fig:afterTsne}.

After the addition of images of shelling defect as training data, the aforementioned image was categorized in the same group as other images of shelling defect, showing that the defect detection model identified the image as similar to other images of shelling defect.

\section{Conclusion and future implications}
In this research, the accuracy of the defect detection model was successfully improved by using CycleGAN to generate images of defective equipment that were indistinguishable from real images and using them as training data. Detecting defects on rails were the subject of this study but we think these findings are applicable to defect detection of various equipment. We plan to conduct further studies with the aim of improving defect detection accuracy using CycleGAN not only for railway tracks but railway equipment in general.


\bibliography{main}

\begin{thebibliography}{1}
\providecommand{\url}[1]{#1}
\csname url@samestyle\endcsname
\providecommand{\newblock}{\relax}
\providecommand{\bibinfo}[2]{#2}
\providecommand{\BIBentrySTDinterwordspacing}{\spaceskip=0pt\relax}
\providecommand{\BIBentryALTinterwordstretchfactor}{4}
\providecommand{\BIBentryALTinterwordspacing}{\spaceskip=\fontdimen2\font plus
\BIBentryALTinterwordstretchfactor\fontdimen3\font minus
  \fontdimen4\font\relax}
\providecommand{\BIBforeignlanguage}[2]{{%
\expandafter\ifx\csname l@#1\endcsname\relax
\typeout{** WARNING: IEEEtran.bst: No hyphenation pattern has been}%
\typeout{** loaded for the language `#1'. Using the pattern for}%
\typeout{** the default language instead.}%
\else
\language=\csname l@#1\endcsname
\fi
#2}}
\providecommand{\BIBdecl}{\relax}
\BIBdecl

\bibitem{jreMonitoring}
R.~Kasai, Y.~Saito, Y.~Komatsu, K.~Ogiso, H.~Yahagi, and T.~Konishi,
  ``{Development of a Track Facility Monitoring Device and Future Prospects for
  the Device},''
  \url{https://www.jreast.co.jp/e/development/tech/pdf_34/tec-34-21-24eng.pdf}.

\bibitem{he2015deep}
K.~He, X.~Zhang, S.~Ren, and J.~Sun, ``Deep residual learning for image
  recognition,'' 2015.

\bibitem{zhu2020unpaired}
J.-Y. Zhu, T.~Park, P.~Isola, and A.~A. Efros, ``Unpaired image-to-image
  translation using cycle-consistent adversarial networks,'' 2020.

\bibitem{Selvaraju_2019}
\BIBentryALTinterwordspacing
R.~R. Selvaraju, M.~Cogswell, A.~Das, R.~Vedantam, D.~Parikh, and D.~Batra,
  ``Grad-cam: Visual explanations from deep networks via gradient-based
  localization,'' \emph{International Journal of Computer Vision}, vol. 128,
  no.~2, p. 336–359, Oct 2019. [Online]. Available:
  \url{http://dx.doi.org/10.1007/s11263-019-01228-7}
\BIBentrySTDinterwordspacing

\bibitem{vanDerMaaten2008}
\BIBentryALTinterwordspacing
L.~van~der Maaten and G.~Hinton, ``Visualizing data using {t-SNE},''
  \emph{Journal of Machine Learning Research}, vol.~9, pp. 2579--2605, 2008.
  [Online]. Available: \url{http://www.jmlr.org/papers/v9/vandermaaten08a.html}
\BIBentrySTDinterwordspacing

\end{thebibliography}

\end{document}